\documentclass{article}

\usepackage{microtype}
\usepackage{graphicx}
\usepackage{subfigure}
\usepackage{booktabs} %

\usepackage{hyperref}

\usepackage[accepted]{icml2024}

\usepackage{amsmath}
\usepackage{amssymb}
\usepackage{mathtools}
\usepackage{amsthm}

\usepackage[capitalize,noabbrev]{cleveref}

\usepackage{breakcites}

\theoremstyle{plain}

\theoremstyle{definition}

\theoremstyle{remark}

\usepackage{times}
\usepackage{latexsym}

\usepackage{tabularx}
\usepackage{multirow}
\usepackage{multicol}
\usepackage{adjustbox}
\usepackage{array}
\usepackage{arydshln}
\usepackage{anyfontsize}

\usepackage[T1]{fontenc}

\usepackage[utf8]{inputenc}

\usepackage{cleveref}
\crefname{figure}{Fig.}{Figs.}
\crefname{table}{Table}{Tables}
\crefname{appendix}{App.}{Apps.}
\crefname{section}{\S}{\S\S}
\crefformat{section}{\S#2#1#3} %
\crefname{equation}{Eq.}{Eqs.}
\crefname{algorithm}{Alg.}{Algs.}
\crefname{algocf}{Alg.}{Algs.}

\usepackage{color}
\usepackage{soul}
\usepackage{xspace}
\usepackage{xparse}

\usepackage[textsize=scriptsize, disable]{todonotes}

\makeatletter
\newcommand*\iftodonotes{\if@todonotes@disabled\expandafter\@secondoftwo\else\expandafter\@firstoftwo\fi}  %
\makeatother

\definecolor{darkblack}{rgb}{0.0,0.0,0.5}
\definecolor{darkgreen}{rgb}{0.0, 0.42, 0.24}
\definecolor{lightgreen}{rgb}{0.52, 0.73, 0.4}
\definecolor{darkgray}{rgb}{0.4,0.4,0.4}
\definecolor{darkblue}{rgb}{0.0,0.0,0.5}
\definecolor{darkpurple}{rgb}{0.5,0.2,0.8}
\definecolor{lightpurple}{rgb}{0.8,0.5,1}

\newcommand{\formalismname}{\textsc{MathWorld}\xspace}
\newcommand{\mathworld}{\formalismname}

\newcommand{\defn}[1]{\textbf{#1}}

\usepackage[group-separator={,},group-minimum-digits={3}]{siunitx}

\newcommand{\lform}[1]{\texttt{#1}\xspace}
\NewDocumentCommand{\transfer}{g}{\lform{transfer}\IfNoValueTF{#1}{}{\!\!\texttt{(#1)}}}
\NewDocumentCommand{\rate}{g}{\lform{rate}\IfNoValueTF{#1}{}{\!\!\texttt{(#1)}}}
\NewDocumentCommand{\comparison}{g}{\lform{comparison}\IfNoValueTF{#1}{}{\!\!\texttt{(#1)}}}
\NewDocumentCommand{\container}{g}{\lform{container}\IfNoValueTF{#1}{}{\!\!\texttt{(#1)}}}

\renewcommand\cite{\citep}	%
\newcommand{\citeposs}[1]{\citeauthor{#1}'s (\citeyear{#1})}

\newcommand{\llama}{LLaMA2\xspace}
\newcommand{\mistral}{Mistral\xspace}
\newcommand{\mixtral}{Mixtral\xspace}
\newcommand{\gpt}{GPT\xspace}

\usepackage{inconsolata}

\icmltitlerunning{Do Language Models Exhibit the Same Cognitive Biases in Problem Solving as Human Learners?}

\begin{document}

\twocolumn[
\icmltitle{Do Language Models Exhibit the Same Cognitive Biases in Problem Solving\\as Human Learners?\looseness=-1}

\icmlsetsymbol{equal}{*}

\begin{icmlauthorlist}
\icmlauthor{Andreas Opedal}{equal,ethz,mpi}
\icmlauthor{Alessandro Stolfo}{equal,ethz}
\icmlauthor{Haruki Shirakami}{ethz}
\icmlauthor{Ying Jiao}{leuven}
\\
\icmlauthor{Ryan Cotterell}{ethz}
\icmlauthor{Bernhard Sch\"olkopf}{ethz,mpi}
\icmlauthor{Abulhair Saparov}{nyu}
\icmlauthor{Mrinmaya Sachan}{ethz}
\end{icmlauthorlist}

\icmlaffiliation{ethz}{ETH Z\"urich}
\icmlaffiliation{leuven}{KU Leuven}
\icmlaffiliation{nyu}{New York University}
\icmlaffiliation{mpi}{Max Planck Institute for Intelligent Systems}

\icmlcorrespondingauthor{Andreas Opedal}{andreas.opedal@inf.ethz.ch}
\icmlcorrespondingauthor{Alessandro Stolfo}{alessandro.stolfo@inf.ethz.ch}

\icmlkeywords{LLMs, Cognitive Science}

\vskip 0.3in
]

\printAffiliationsAndNotice{\icmlEqualContribution} %

\begin{abstract}
There is increasing interest in employing large language models (LLMs) as cognitive models.
For such purposes, it is central to understand which properties of human cognition are well-modeled by LLMs, and which are not. In this work, we study the biases of LLMs in relation to those known in children when solving arithmetic word problems. Surveying the learning science literature, we posit that the problem-solving process can be split into three distinct steps: text comprehension, solution planning and solution execution. We construct tests for each one in order to understand whether current LLMs display the same cognitive biases as children in these steps. 
We generate a novel set of word problems for each of these tests, using a neuro-symbolic approach that enables fine-grained control over the problem features.
We find evidence that LLMs, with and without instruction-tuning, exhibit human-like biases in both the text-comprehension and the solution-planning steps of the solving process, but not in the final step,
in which the arithmetic expressions are executed to obtain the answer.

\vspace{.11em}
\hspace{1.25em}\includegraphics[width=1.25em,height=1.25em]{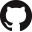}{\hspace{.75em}\parbox{\dimexpr\linewidth-2\fboxsep-2\fboxrule}{\url{https://github.com/eth-lre/solving-biases}}}
\end{abstract}
\vspace{-22pt}
\section{Introduction}

There is active discussion around large pretrained language models (LLMs) as plausible cognitive models \citep{mahowald-ivanova-2023}, e.g., in terms of language acquisition \citep{warstadt2022artificial}, decision making \citep{aher-using-2023} and political orientation \citep{argyle_busby_fulda_gubler_rytting_wingate_2023}.
In the context of education, cognitive modeling enables the study of human learning without the high cost of data collection from human subjects, which can lead to a better understanding of human learning and, therefore, improved learning outcomes
\citep{vanlehn-applications-1994}.
Several recent articles have already employed LLMs as models of students \citep{macina-etal-2023-mathdial, nguyen2023large}. However, for such modeling to be meaningful, it is imperative that the student model is consistent with actual student behavior. 
Yet, that is not always the case: Many existing student models fail to validate faithfulness to realistic classroom scenarios \citep{Kaser-Alexandron-2023}. 
Importantly, an LLM that models the problem-solving process of children should also make similar mistakes as children, i.e., it should mimic the cognitive biases that are salient in children during problem-solving.
Given that LLMs may be trained predominantly on data generated by \emph{adults}, it is not obvious that they would exhibit \emph{child-like} behavior.\looseness=-1

\begin{figure}[t]
    \centering
    \includegraphics[width=0.49\textwidth]{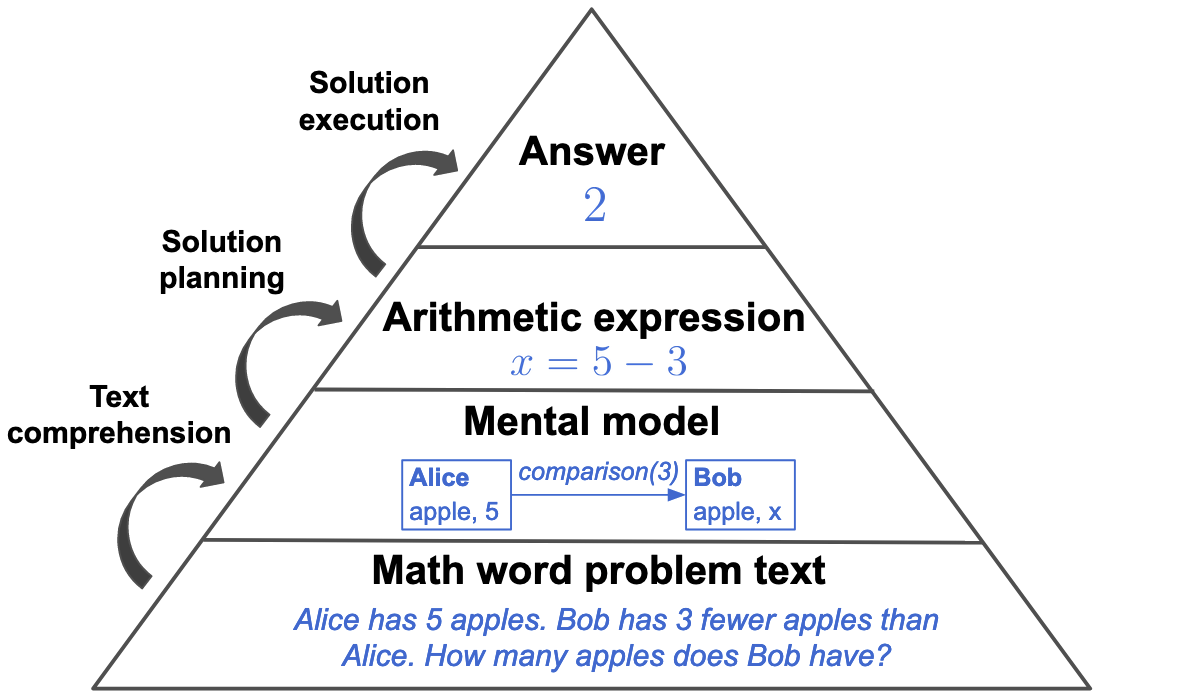}
    \vspace{-16pt}
    \caption{A three-step model of the cognitive process involved in solving math word problems.
    We study whether LLMs exhibit similar biases as human children along each step of this process. 
    }
    \label{fig:solving-pyramid}
    \vspace{-19pt}
\end{figure}

In this paper, we study whether LLMs are subject to similar biases as children when solving arithmetic math word problems.\footnote{Related studies have compared LLM biases to human ones on the task of syllogistic reasoning \citep{ando-etal-2023-evaluating, dasgupta2023language, eisape2023systematic}. These are discussed in \cref{sec:related-work}.} 
These problems are interesting because they are conceptually simple,
and yet, require several distinct skills to solve \citep{stern_1993}. A learner 
needs to understand the situation described, relate it to arithmetic equations, and perform the required computations, as \cref{fig:solving-pyramid} illustrates.
By \citeposs{piaget1936} view on cognitive development, a problem might be difficult for a child due to insufficient development in any one of these skills.
Much is known about what makes a word problem difficult for humans; we ask whether the same relative difficulties apply to LLMs.

To answer this question, we 
construct tests that are grounded in the extensive literature on word problem solving by children,\footnote{This comment refers to studies performed on children, but we note that some of the biases considered seem to be present in adults as well (albeit with weaker effects). See \citet{Jaffe-bolger-2023-survey} for a recent review on word problem performance independent of age.\looseness=-1 } and perform them on a suite of currently well-known LLMs.
Specifically, 
each test varies a problem feature for which an effect on child performance has been established in the literature, e.g., the manner in which a particular mathematical operation is expressed in text, while controlling for other features. 
We create new English problems specifically for these tests, by developing a generation pipeline based on a semantic formalism over math word problems \citep{opedal-etal-2023-world}.
Our generation pipeline admits a family of standard arithmetic word problems, while controlling not only for numerical features, but structural (e.g., entity relationships) and linguistic ones (e.g., sentence structure) as well.\looseness=-1

We test three cognitive biases, each one associated with a separate step of the solving process (which are illustrated by arrows in \cref{fig:solving-pyramid}).
The first test targets \defn{consistency bias}: A problem text is easier to comprehend if the relational keyword verbally suggests the operation that is required to solve the problem \citep{Lewis1987StudentsMO}. The second test targets what we call \defn{transfer vs comparison bias}, 
that problems with a static comparison relationship are more difficult for children than problems with a dynamic change of state, even when they involve the same arithmetic expressions \citep{riley-1983-development}. The third test targets the \defn{carry effect}, i.e., that arithmetic computations are more difficult if they entail moving some digit to a column of more significant digits \citep{HITCH1978302}.\looseness=-1

We find that LLMs indeed exhibit some biases that mirror those observed in children. Our experiments with both base and instruction-tuned models---specifically, \llama \cite{touvron2023llama}, \mistral \cite{jiang2023mistral}, \mixtral \cite{jiang2024mixtral}, \gpt-3.5 Turbo and \gpt-4 Turbo \citep{openai2024gpt4}---reveal that almost all models show significant effects of consistency bias and transfer vs comparison bias, 
like child learners. Most of these effects are further strengthened by using chain-of-thought prompting \citep{wei_chain_of_thought}.
However, we do \emph{not} observe a
carry effect bias in the solution execution step.
These results contribute to our understanding of the capabilities and limitations of LLMs as cognitive models, particularly in the context of cognitive development research and educational applications. \looseness=-1

\section{Cognitive Modeling of the Solving Process
}
\label{sec:background-cogsci}
This section discusses the cognitive process that is involved in solving math word problems. 
We first introduce our conceptualization that is illustrated in \cref{fig:solving-pyramid}, which we then motivate by summarizing relevant literature.

\paragraph{Our conceptualization.} We are interested in identifying \emph{when} LLMs are likely to exhibit human-like biases, and therefore, require a holistic analysis of the human problem-solving process.
Our conceptualization, illustrated in \cref{fig:solving-pyramid}, includes four representational levels of the math word problem, along with the skills associated with transitioning from one level to the next. We assume that a child goes through the following procedure when posed with a math word problem:
First, they form a \defn{mental model} 
of the mathematical relationships expressed in that problem (text comprehension). Next, they distill that mental model into a sequence of \defn{arithmetic expressions} representing the step-by-step reasoning process to find the solution (solution planning) and, finally, calculate the \defn{answer} from those expressions (solution execution). The representational levels will be formalized in \cref{sec:mathworld}.\looseness=-1

\paragraph{Background.} Children tend to experience greater difficulty when posed with arithmetic math word problems compared to the same problems formulated solely as arithmetic expressions; see, e.g., \citet{Jaffe-bolger-2023-survey}. 
This suggests that arithmetic computation skill alone is not sufficient to successfully solve math word problems. In order to distinguish the different skills that are involved, past work has represented math word problems along similar levels as we do above \citep{nesher-teubal-1975-verbal}: (i) problem text, (ii) underlying mathematical relationships, and (iii) arithmetic expressions.
Solving a problem, then, involves transitioning from level (i) to a final answer, possibly through levels (ii) and (iii), with each transition requiring a separate skill.\footnote{Not everyone uses all three levels in their solving process. \citet{Hegarty1995ComprehensionOA} find evidence that unsuccessful human problem solvers often opt for short-cut strategies that rely on surface-level features of the problem text
(thus, by our conceptualization, 
moving directly from text to arithmetic expressions), whereas successful solvers are more likely to make use of mental models. \label{fn:shortcuts}
}

There is much research on which factors best explain problem difficulty. \citeposs{riley-1983-development} model of the problem-solving process emphasized the degree of complexity at level (ii)
as the leading factor underlying performance, motivated by empirical evidence that some arithmetic concepts are harder for children to understand than others. However, their model does not account for \emph{how} the mathematical relationships are derived from the problem text \citep{CUMMINS1988405}. This part is significant as well, as several studies have found that altering the linguistic form of a problem \emph{without} changing the underlying mathematical relationships can have drastic effects on performance (\citealp{Hudson1983CorrespondencesAN, Lewis1987StudentsMO}; \emph{inter alia}) in both children and adults. This part of the process is encompassed by the models of \citet{briars-larkin-1984} and \citet{Kintsch1985-KINUAS}.
None of these models give an explicit account of the complexity of the arithmetic expression, however, which also has significant influence on performance
\citep{daroczy-et-al-2015-word}. %
\looseness=-1

\section{
Human Biases in Word Problems
}
\label{sec:hypotheses}

In this section we discuss the particular factors that influence performance of human children (i.e., cognitive biases) 
which we study in LLMs (\cref{sec:experiments}). 
Each bias is reflected by a variation in 
a specific level of \cref{fig:solving-pyramid}. We study one bias for each of the three levels that precedes the answer. 

\paragraph{Problem text: Consistency bias.} 
Given the premise ``Alice has 5 apples'' and a question querying the (smaller) number of apples of another agent ``Bob'', an additive comparison statement
between the two agents could take either of the following forms:

\vspace{0.6em}
\begin{minipage}{19em}
    (1) Bob has 3 fewer apples than Alice. \\
    (2) Alice has 3 more apples than Bob.
\end{minipage}
\vspace{0.3em}

\noindent Here, (1) represents a \emph{consistent} statement because the relational keyword (``fewer'') suggests the operation that is indeed needed to compute Bob's quantity (subtraction). Conversely, (2) is an \emph{inconsistent} statement because the relational keyword (``more'') suggests a different operation (``addition''). 
Note that these two statements express the same comparison relationship, so the difference lies only in the problem text.
Problems with an inconsistent statement are more difficult for children to solve than consistent ones \citep{nesher-teubal-1975-verbal, stern_1993}. This has been hypothesized to be the case due to inconsistent statements requiring an additional, error-prone, deduction step: to rearrange the relational statement to be in the preferred consistent form \citep{Lewis1987StudentsMO}.\looseness=-1

\paragraph{Mental model: Transfer vs comparison bias.
} 
Irrespective of which relational keyword is used, comparison-type problems tend to be more difficult for children than other types of arithmetic concepts \citep{riley-1983-development}. In particular, grade school children display a significant difference in performance between comparison problems and transfer problems. Consider the same premise as above but with a slightly different continuation:

\vspace{0.6em}
\begin{minipage}{19em}
    Alice has 5 apples. Alice gave 3 apples to Bob. How many apples does Alice have?
\end{minipage}
\vspace{0.3em}

\noindent This is a transfer problem (often
 called a \emph{change} elsewhere;
\citealp{Nesher1982TheDO}), concerning a change of state of some quantity. 
It has the same arithmetic expression as the comparison problems above (although with another mental model), but is easier for young children to solve. In analyzing their solution strategies,
it has been found that
comparison problems require a number-matching type strategy that appears to be more sophisticated than the counting-type strategies that are often sufficient for solving transfer problems \citep{riley-1983-development, carpenter-moser-1984-acquisition}.

\paragraph{Arithmetic expressions: The carry effect.} 
Beyond the text and mental model, it is natural that the particular numbers used in a problem will have an effect on a child's performance \citep{daroczy-et-al-2015-word}. Consider the same problem(s) as above, but with a different number given in the premise:\looseness=-1

\vspace{0.6em}
\begin{minipage}{19em}
    Alice has 22 apples. Bob has 3 fewer apples than Alice. How many apples does Bob have?
\end{minipage}
\vspace{0.3em}

\noindent The problem now has the arithmetic expression $22-3$, which involves an arithmetic \emph{carry}, which is also called a borrowing in the case of subtraction.
A carry is a digit that is transferred from one column of digits to another as the result of an arithmetic computation. In this subtraction computation, a unit carry of $1$ is transferred from the column of units to the column of tens in order to make the answer $19$. 
The previous expression ($5-3$) did not have a carry, which is easier for children \citep{HITCH1978302, ASHCRAFT1992301}.
The presence of a carry introduces an additional subroutine from the standard sequence of operations, which places a higher load on working memory \citep{Fuerst2000carry}.\footnote{The particular example numbers given here are small enough for children to likely be relying on retrieval from some long-term memory store instead of algorithmic computation \citep{KOSHMIDER199153}, which could erase the effect of carry. We account for this in our experiments (\cref{sec:numerical-tests}) by using larger number ranges.\label{fn:memory}}\looseness=-1 

\section{Problem Generation Method
}\label{sec:generation}

Our experiments on LLMs with respect to the biases just discussed (\cref{sec:experiments}) rely on data generated for the sole purpose of our study. By not using problems from public datasets, previous work or other existing sources it becomes unlikely that our data has been used for training of the models, an increasingly common issue \citep{dodge-etal-2021-documenting, elazar2023s}.\looseness=-1

This section gives the details of our data generation pipeline, which provides control over features across all levels of \cref{fig:solving-pyramid}. 
In \cref{sec:mathworld}, we operationalize \cref{fig:solving-pyramid}, giving definitions related to the mental model level and other aspects of the process. Using these definitions, we then explain our generation pipeline in \cref{sec:generation-pipeline}.\looseness=-1

\subsection{
Operationalizing \cref{fig:solving-pyramid}
}
\label{sec:mathworld}
The mental model level is operationalized using the formalism from \citet{opedal-etal-2023-world}, called \mathworld. \mathworld annotates each math word problem with a logical representation, which captures the mathematical relationships between the entities described in text. 
Each \defn{entity} has a non-negative integer \defn{quantity}.
Optionally, there may be additional information associated with an entity---namely, an \defn{agent} who possesses the entity, and a \defn{unit} and an \defn{attribute} which enrich the description of it. 
The five bold items in the two preceding sentences are referred to as \defn{properties}.
The arithmetic relationships are classified according to \defn{concepts}; we use transfer, comparison (additive and multiplicative), and rate in this work. We gave examples of the transfer and comparison concepts in \cref{sec:hypotheses}.

When discussing data generation (\cref{sec:generation-pipeline}) and the experiments the data is used for (\cref{sec:experiments}), we will rely on the following definitions:
A \defn{predicate} is a symbol that represents either an arithmetic concept, or possession of an entity (in that case the predicate is ``{\small \container}''). Each predicate takes a set of properties as arguments.\footnote{We enforce all quantities that are associated with predicates to be explicit numbers. Note that this places a constraint on the format of the problems: The number associated with a mathematical relationship is never an intermediate result, but is always given in text.\looseness=-1 
}
A predicate instantiated with properties is called a \defn{logical form}, and represents the semantics of a given sentence in a problem.
See \cref{table:lf_examples} for examples of logical forms for all predicates we use.
The \defn{mental model} of a problem is 
a sequence of logical forms (separated by a ``$\circ$'' symbol) for each sentence in that problem (in the same order), representing its semantics. 
In \cref{fig:solving-pyramid}, we gave a mental model example in graphical format; its equivalent sequential format is {\small \texttt{container(Alice, 5, apple)} $\circ$  \texttt{comparison(}$+$\texttt{, Alice, Bob, 3, apple)}}. 
The problem text in \cref{fig:solving-pyramid} is \defn{faithful} to this mental model, in the sense that the mental model represents the semantics of that text under the \mathworld formalism.
We refer to the \defn{structure} of a problem as a mental model in which the property values are replaced by unique placeholders. The structure associated with the previous example is {\small \texttt{container([agent1], [n1], [entity1])} $\circ$  \texttt{comparison(}$+$\texttt{, [agent1], [agent2], [n2], [entity1])}}.\looseness=-1

\newcolumntype{Z}{>{\raggedright\arraybackslash\hsize=.19\hsize}X}
\newcolumntype{K}{>{\raggedright\arraybackslash\hsize=.35\hsize}X}
\newcolumntype{Q}{>{\raggedright\arraybackslash\hsize=.46\hsize}X}

\makeatletter
\def\adl@drawiv#1#2#3{%
        \hskip.5\tabcolsep
        \xleaders#3{#2.5\@tempdimb #1{1}#2.5\@tempdimb}%
                #2\z@ plus1fil minus1fil\relax
        \hskip.5\tabcolsep}
\newcommand{\cdashlinelr}[1]{%
  \noalign{\vskip\aboverulesep
           \global\let\@dashdrawstore\adl@draw
           \global\let\adl@draw\adl@drawiv}
  \cdashline{#1}
  \noalign{\global\let\adl@draw\@dashdrawstore
           \vskip\belowrulesep}}
\makeatother

\newcommand{\textchange}[1]{{\normalsize \textit{#1}}}

\begin{table}
\resizebox{\columnwidth}{!}{
    \begin{tabularx}{1.4\columnwidth}{Z K Q}
    \toprule[0.1em]
    \multicolumn{2}{c}{\textbf{Logical Form}} & \multirow{2}{\hsize}{\textbf{Example Sentences}} \\\cmidrule{1-2}
    \textbf{Predicate} & \textbf{Properties} & \\
    \midrule
    \multirow{5}{*}{\texttt{container}} & \texttt{agent=}\textit{Alice} & \multirow{2.5}{\hsize}{\textchange{Alice has 5 kilograms of red apples.}} \\
    & \texttt{quantity=}5 & \\ %
    & \texttt{entity=}\textit{apple} & \multirow{3}{\hsize}{\textchange{Alice owns 5 kilograms of red apples.}}\\
    & \texttt{attribute=}\textit{red} & \\
    & \texttt{unit=}kg & \\\cmidrule{1-3}
    \multirow{5}{*}{\texttt{comparison}} & \texttt{type=$+$} &  \multirow{2.5}{\hsize}{\textchange{Bob has 3 fewer apples than Alice.}} \\
    &  \texttt{agentA=}\textit{Alice} & \\ %
    &  \texttt{agentB=}\textit{Bob} &  \multirow{3}{\hsize}{\textchange{Alice has 3 more apples than Bob.}} \\
    &  \texttt{quantity=}3 & \\
    &  \texttt{entity=}\textit{apple} & \\\cmidrule{1-3}
    \multirow{4}{*}{\texttt{transfer}} & \texttt{receiver\_agent=}\textit{Bob} & \multirow{2}{\hsize}{\textchange{Alice gave Bob 3 apples.}} \\
    &  \texttt{sender\_agent=}\textit{Alice} & \\ %
    &  \texttt{quantity=}3 & \multirow{2}{\hsize}{\textchange{Bob got 3 more apples from Alice.}}\\
    &  \texttt{entity=}\textit{apple} & \\\cmidrule{1-3}
    \multirow{4}{*}{\texttt{rate}} & \texttt{agent=}\textit{Alice} & \multirow{2}{\hsize}{\textchange{Each of Alice’s baskets holds 4 apples.}} \\
    &  \texttt{quantity=}4 & \\ %
    &  \texttt{entityA=}\textit{apple} & \multirow{2}{\hsize}{\textchange{Every basket that Alice has contains 4 apples.}}\\
    &  \texttt{entityB=}\textit{basket} & \\
 \bottomrule[0.1em]
\end{tabularx}
}
\vspace{-6pt}
    \caption{Examples of \mathworld logical forms. A logical form consists of a predicate and a set of properties. Note that the \emph{attribute} and \emph{unit} properties are optional, and only the first logical form has non-null values for them (``red'' and ``kg'', respectively). Each logical form is given with two of the viable templated sentences from our generation pipeline (\cref{sec:generation-pipeline}), the semantics of which is represented by the logical form.
    }
    \label{table:lf_examples}
    \vspace{-10pt}
\end{table}

Finally, we formalize the arithmetic expression level of \cref{fig:solving-pyramid}.
Every concept-based predicate corresponds to an equation $x = y \odot z$, with $\odot \in \{+,-,\times,\div\}$ and $x,y,z \in \mathbb{Z}_{\geq 0} \cup V$ where $V$ is a set of variable symbols. 
We require that $\exists! v \in \{x,y,z\}\colon v\in V$.
We refer to the deductive inference step taken to solve that equation as a \defn{reasoning step}, and its output (i.e., the value of the variable)
as an \textbf{intermediate result}. The arithmetic expression level consists of a sequence of such reasoning steps, which is a proof of the answer of the problem (or solution). Any fact from the mental model is an axiom that can be used in the solution proof.
This work focuses exclusively on \defn{linear} problems, in which every reasoning step has \emph{at most} one non-axiom premise.

\paragraph{Plausibility of the mental model framework.
}
A mental model theory over some reasoning domain must be able to adequately explain the relative difficulty across different types of reasoning problems \citep{johnson-laird-1983}. Our \mathworld-based operationalization emphasizes arithmetic concepts and their relational structure as the key features that explain errors at the mental model level, which is in line with existing theories on word problems \citep{riley-1983-development, briars-larkin-1984, Kintsch1985-KINUAS}. Our schemata for the logical forms extend the problem schemata from \citet{riley-1983-development}, specifically, in terms of the breadth of concepts and properties supported.

\subsection{Generation Pipeline}
\label{sec:generation-pipeline}
\begin{figure*}[ht!]
    \centering
    \includegraphics[width=0.98\textwidth]{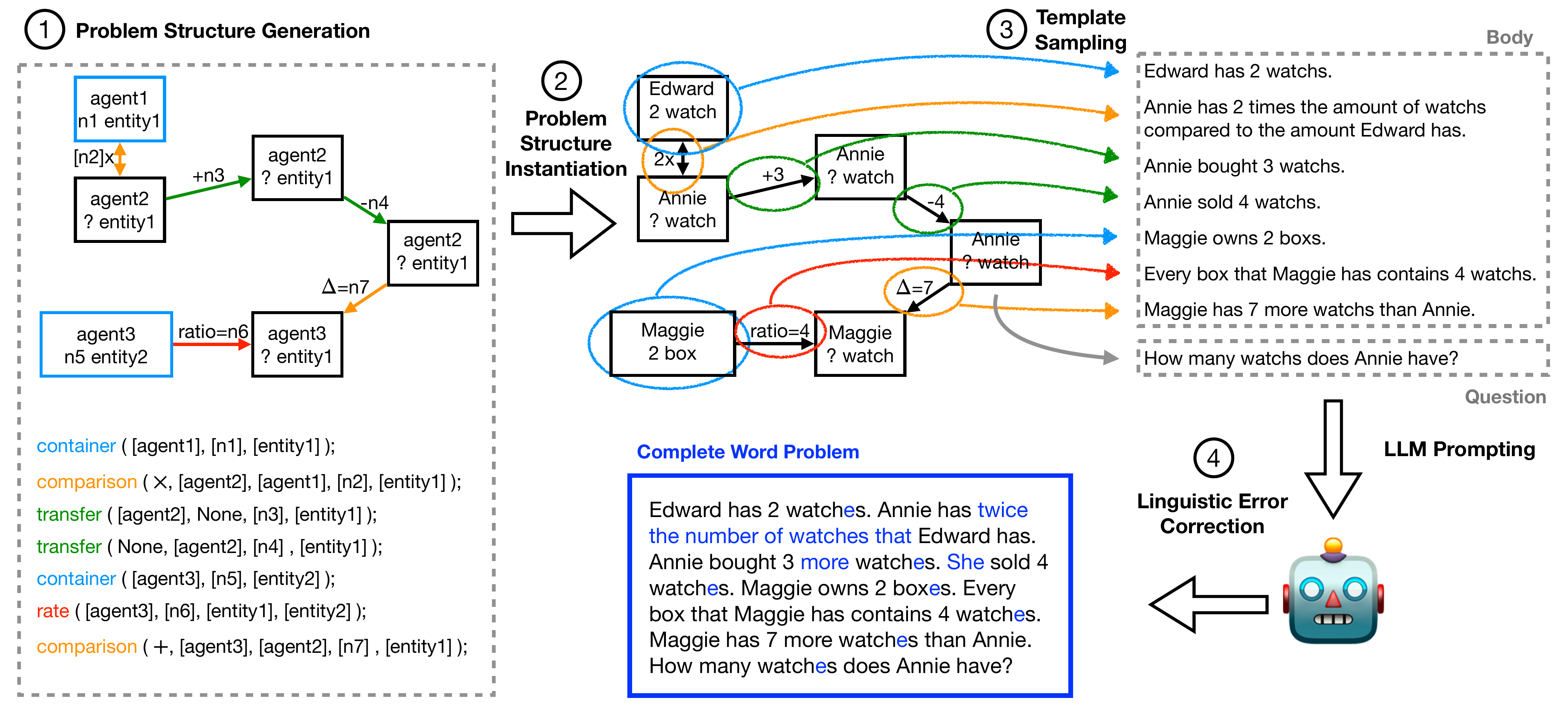}
    \vspace{-2pt}
    \caption{Overview of our generation pipeline with an example problem.
    (1) We start by generating the problem structure. %
    The alignment between the graphical and sequential formats of the structure is illustrated by color coding, and the \textbf{black} containment boxes in the graph represent intermediate results.
    (2) The properties of those structures are then instantiated with values, resulting in a mental model. (3) Next, we sample a templated sentence for each of the logical forms in the mental model, and concatenate them in the ordering of the logical forms. (4) Finally, we correct errors and awkward phrasings by prompting an instruction-tuned LLM (the corrections are highlighted in {\color{blue} \textbf{blue}}). This particular example includes all predicate types that we use in \cref{sec:experiments}.}
    \label{fig:generation-pipeline}
    \vspace{-12.5pt}
\end{figure*}

We propose a simple problem generation pipeline, described in this section and later applied in \cref{sec:experiments}. It proceeds in four steps:
(i) sampling the (linear) problem structure, (ii) obtaining a mental model by instantiating the structure with properties, (iii) transforming the mental model into templated natural language, and, finally, (iv) correcting linguistic errors and awkward phrasings in the templated text using an instruction-tuned LLM.
\cref{fig:generation-pipeline} illustrates the full pipeline.
\cref{sec:related-methods} discusses related approaches.

\paragraph{(i) Problem structure generation.
}
Our pipeline supports sequences of predicates under the following regular language, which is sufficient for the hypotheses we are testing:
\vspace{1mm}
\resizebox{\linewidth}{!}{
  \begin{minipage}{\linewidth}
\begin{align*}\
    & \small \container \circ \underbrace{(\texttt{concept} \mid \container)}_{1-N\ \text{times}}; 
\end{align*}
\end{minipage}
}

where {$\small \texttt{concept} = \comparison \mid \rate \mid \transfer$}.
Each predicate in the sequence corresponds to an axiom from a distinct sentence in the problem. The particular set of predicate sequences (i.e., class of structures) from which we sample is test dependent (see \cref{sec:experiments}).
Such a set could, for instance, be the class of all linear reasoning structures with at most $N{=}5$ steps and only transfer concepts.
We generate a problem as follows: First, we sample the number of reasoning steps $n$ uniformly at random from the set $\{1, \ldots, N\}$. 
Next, we sample the predicates; each choice is sampled uniformly at random.
Since this first step of the pipeline generates \emph{structures}, the predicates all have associated unique placeholders in place of properties, e.g., {\small \lform{agent2}}, {\small \lform{entity1}}. 
We only introduce new entity placeholders in rate logical forms; see \cref{table:lf_examples}, in which {\small \rate} is the only predicate that takes two entity properties.
We determine uniformly at random whether an entity is paired with an attribute, a unit, or neither.
The instantiation of agent placeholders is test specific. Finally, the answer of a problem is always set as the intermediate result corresponding to the last logical form in the ordering, which is unique. See the box on the left-hand side of \cref{fig:generation-pipeline} for an example structure generated after this step.

\paragraph{(ii) Problem structure instantiation.}
Next, the problem structure is instantiated with properties, yielding a mental model. We use a handwritten vocabulary for each of the lexical properties (entity, agent, unit and attribute) and sample instantiations of these properties from those vocabularies uniformly at random.
The numerical quantities are instantiated by sampling a set of numbers uniformly at random from within a fixed range, which is $2$--$20$ for the experiments in \cref{sec:consistency-hypothesis}-\ref{sec:concept-test} and $100$--$999$ for the experiments in \cref{sec:numerical-tests}.\footnote{We omit 0 and 1 from the first number range in order to avoid unnatural phrases like ``\textit{Bob has 1 times as many apples as Alice.}'' The range for the experiments in \cref{sec:numerical-tests} contains larger numbers for reasons given in \cref{fn:memory}.} Then, we enumerate the logical forms and accordingly compute intermediate results for each reasoning step, making sure that the intermediate quantities fall in the range $0$--$999$. If not, a new set of numbers is sampled and the procedure is repeated from the beginning. 
This naive procedure is sufficiently fast for our purposes.
Empirically, we observed an average runtime of $\approx$4 ms to generate a numerical instantiation for a problem.\looseness=-1

\paragraph{(iii) Template sampling.}
The mental model is then converted to natural language using templated text.
Specifically, for each of the predicates 
we construct a set of templates that represent natural language adhering to that predicate. For instance, {\small \transfer{Annie, None, 3, watch}} is converted to ``[Annie]\textit{ bought} [3] [watch]\textit{s}'' in \cref{fig:generation-pipeline}; see \cref{table:lf_examples} for additional examples.
The templates are handcrafted.
We sample one template uniformly at random for each predicate in the mental model. We also create and sample interrogative templates for the questions, which always query the intermediate result derived from the \emph{last} predicate. Finally, we concatenate the sentences to obtain the full problem text.
This step enables control over the linguistic form of the sentences in the problem text, which will be important for our test related to text comprehension in \cref{sec:linguistic-tests}.
Moreover, since the faithfulness of the templated text is guaranteed by manual design, the procedure up to this point ensures that the text is faithful to the mental model.\looseness=-1

\paragraph{(iv) Linguistic error correction.}
However, templated texts occasionally induce spelling mistakes and awkward phrasings. In the example shown in \cref{fig:generation-pipeline}, the entity ``watch'' is inserted into the template to make ``watchs''. Inspired by the demonstrated success of zero-shot grammatical error correction \citep{kwon-etal-2023-beyond, loem-etal-2023-exploring}, we use an instruction-tuned language model \citep{ouyang2022training}, GPT-3.5 Turbo,
to correct such errors.\footnote{Interestingly, we found that GPT-3.5 Turbo sometimes would transform inconsistent formulations of comparison-type relationships into semantically equivalent consistent ones, which already indicates presence of a bias towards consistent problems. Such erroneous transformations were filtered out (see \cref{app:error-correction}).
}
We write a short instructive prompt and have the model generate (with greedy decoding) a corrected problem text
conditioned on that prompt together with the particular templated text we want to correct. The prompt instructs the model to be conservative, i.e., to only correct linguistic errors and awkward phrasings.
We provide the exact prompt used and additional generation details in \cref{app:error-correction}.
This step could, in principle, be generalized to perform less strict forms of paraphrasing. There is then a trade-off between faithfulness and control on the one hand and linguistic variability and naturalness on the other, which can be tuned using different prompts and decoding methods. The present study prioritizes control and faithfulness, but alternative prioritizations could be used in future studies that employ our method.

\paragraph{Evaluating data quality.} The generated problem texts must be faithful to the mental models from which they were generated, so we perform manual evaluation of the data to assess that such is the case. We follow a generic procedure and perform it for each of the datasets generated for the experiments in \cref{sec:linguistic-tests}-\ref{sec:numerical-tests}. The procedure is iterated until we achieve satisfactory quality. The final error-rate estimates were $0\%$ for all three datasets. Details are given in \cref{sec:data-quality}.

\section{Experiments}\label{sec:child-learning}\label{sec:experiments}
We generate data to perform tests on whether LLMs exhibit child-like biases using the pipeline discussed above. We aim to identify where in the process in \cref{fig:solving-pyramid} those biases emerge. We therefore split our tests according to the level (and associated skill) they target: problem text and text comprehension (\cref{sec:linguistic-tests}), mental model and solution planning (\cref{sec:structural-tests}), and arithmetic expressions and solution execution (\cref{sec:numerical-tests}).
First, we discuss the general experimental setup (\cref{sec:experiment-setup}).\looseness=-1

\subsection{Experimental Setup}
\label{sec:experiment-setup}
We base our experiments on the problem features discussed in \cref{sec:hypotheses} that have been found to have an effect on child performance in solving word problems. Specifically, given such a feature $X$, we want to know whether $X$ has a causal effect on the performance of LLMs. Our generation pipeline enables exact matching of the data: We generate problems in pairs, where the two problems differ only in the value of $X$. Using this data, we estimate the \emph{conditional average treatment effect} \cite[CATE;][]{imbens-rubin-causal-2015}
\begin{equation}
    \mathbb{E}[Y(x) - Y(x') \mid Z],
\end{equation}
where $Y$ is $1$ if the model's prediction is correct and $0$ otherwise,
$x$ and $x'$ are two distinct values of the treatment variable $X$, and $Z$ is the subgroup of data generated through our pipeline for a specific test. The two values $x$ and $x'$ are defined such that positive CATEs are consistent with human biases.
We refer to \citet{feder-etal-2022-causal}
for further reading on causality-based methods for NLP.

\newcolumntype{Y}{>{\centering\arraybackslash}X}
\begin{table*}\centering
\resizebox{0.95\textwidth}{!}{
    \begin{tabularx}{1.35\textwidth}{l l *{4}{Y} c *{4}{Y} c *{4}{Y}}
    \toprule[0.1em]
     \multirow{4}{*}{\textbf{Mode}} &
     \multirow{4}{*}{\textbf{Model}} &
    \multicolumn{4}{c}{\textbf{Consistency bias (\cref{sec:consistency-hypothesis})}} & &
    \multicolumn{4}{c}{\textbf{Transfer vs comparison bias (\cref{sec:concept-test})}} & & 
    \multicolumn{4}{c}{\textbf{Carry effect (\cref{sec:numerical-tests})}} \\\cmidrule{3-6}\cmidrule{8-11}\cmidrule{13-16}
     &  & \multicolumn{3}{c}{\textbf{Accuracy (\%)}}& \multirow{2}{*}{$p$-value} & &
     \multicolumn{3}{c}{\textbf{Accuracy (\%)}}& \multirow{2}{*}{$p$-value} & &
     \multicolumn{3}{c}{\textbf{Accuracy (\%)}}& \multirow{2}{*}{$p$-value} \\\cmidrule{3-5}\cmidrule{8-10}\cmidrule{13-15}
      & & \textbf{Co} & \textbf{InCo}  &  CATE & &
     & \textbf{T} & \textbf{C}  &  CATE & & 
     & \textbf{NCa} & \textbf{Ca}   &  CATE &\\
    \midrule
     \multirow{11}{*}{Direct} & \llama 7B 	 & 9.6 & 4.8 & \textbf{4.8} & <0.001 &	 & 21.8 & 13.0 & \textbf{8.8} & <0.001 &	 & 64.8 & 60.0 & \textbf{4.8} & 0.009  \\
                        & \llama 13B 	 & 17.2 & 14.0 & \textbf{3.2} & 0.006 &	 & 28.6 & 20.0 & \textbf{8.6} & <0.001 &	 & 72.2 & 67.2 & 5.0 & 0.030  \\
                        & \llama 70B  & 24.0 & 16.2 & \textbf{7.8} & <0.001 &	& 45.4 & 26.8 & \textbf{18.6} & <0.001 &  &  95.2 &	96.2 & 1.0 & 0.380 \\
                        & \mistral 7B 	 & 17.8 & 12.0 & \textbf{5.8} & <0.001 &	 & 34.0 & 20.4 & \textbf{13.6} & <0.001 &	 & 72.4 & 72.0 & 0.4 & 0.835  \\
                        & \mixtral 8x7B 	 & 23.0 & 17.0 & \textbf{6.0} & <0.001 &	 & 42.2 & 30.4 & \textbf{11.8} & <0.001 &	 & 95.4 & 93.6 & 1.8 & 0.117  \\
                             \cmidrule{2-16} 
                    & \llama 7B Chat 	 & 14.2 & 10.8 & \textbf{3.4} & 0.009 &	 & 20.2 & 15.8 & \textbf{4.4} & 0.005 &	 & 61.2 & 54.2 & 7.0 & 0.012  \\
                    & \llama 13B Chat 	 & 16.4 & 11.8 & \textbf{4.6} & <0.001 &	 & 25.4 & 18.2 & \textbf{7.2} & <0.001 &	 & 65.6 & 59.6 & 6.0 & 0.018  \\
                    & \llama 70B Chat 	 & 16.4 & 14.8 & 1.6 & 0.158 &	 & 32.4 & 20.0 & \textbf{12.4} & <0.001 &	 & 96.4 & 97.0 & -0.6 & 0.578  \\
                    & \mistral 7B Instr. 	 & 17.6 & 14.2 & \textbf{3.4} & 0.008 &	 & 28.0 & 21.8 & \textbf{6.2} & <0.001 &	 & 78.0 & 78.6 & -0.6 & 0.802  \\
                    & \mixtral 8x7B Instr. 	 & 23.4 & 21.8 & 1.6 & 0.195 &	 & 42.6 & 28.0 & \textbf{14.6} & <0.001 &	 & 95.8 & 96.4 & -0.6 & 0.578  \\
                    & \gpt-3.5 Turbo &  32.2 & 22.8 & \textbf{9.4} & <0.001 &	 &  61.0 & 33.4 & \textbf{27.6} & <0.001 &	 &  99.6 & 99.4 & 0.2 & 0.320 \\
      \midrule 
      \multirow{12}{*}{CoT} & \llama 7B 	 & 16.4 & 6.0 & \textbf{10.4} & <0.001 &	 & 18.8 & 13.6 & \textbf{5.2} & 0.009 &	 & 33.2 & 38.8 & -5.6 & 0.006  \\
                            & \llama 13B 	 & 30.2 & 8.6 & \textbf{21.6} & <0.001 &	 & 37.8 & 13.2 & \textbf{24.6} & <0.001 &	 & 33.8 & 33.4 & 0.4 & 0.833  \\
                            & \llama 70B  &  40.2 & 24.0 & \textbf{16.2} & <0.001 &	 &  63.8 & 33.0 & \textbf{30.8} & <0.001 &	 &  68.6 & 67.6  & 1.0 & 0.850 \\
                            & \mistral 7B 	 & 36.4 & 16.8 & \textbf{19.6} & <0.001 &	 & 49.8 & 58.8 &\textbf{ -9.0} & 0.004 &	 & 73.2 & 71.0 & 2.2 & 0.283  \\
                            & \mixtral 8x7B 	 & 62.4 & 42.2 & \textbf{20.2} & <0.001 &	 & 68.6 & 65.0 & 3.6 & 0.206 &	 & 79.8 & 79.8 & 0.0 & 1.000  \\
                           \cmidrule{2-16} 
                            & \llama 7B Chat 	 & 66.8 & 38.6 & \textbf{28.2} & <0.001 &	 & 69.6 & 40.8 & \textbf{28.8} & <0.001 &	 & 72.4 & 71.0 & 1.4 & 0.514  \\
                        & \llama 13B Chat 	 & 67.0 & 28.6 & \textbf{38.4} & <0.001 &	 & 79.4 & 48.0 & \textbf{31.4} & <0.001 &	 & 73.8 & 78.6 & -4.8 & 0.017  \\
                        & \llama 70B Chat 	 & 82.8 & 61.4 & \textbf{21.4} & <0.001 &	 & 99.0 & 76.2 & \textbf{22.8} & <0.001 &	 & 97.0 & 95.8 & 1.2 & 0.180  \\
                        & \mistral 7B Instr. 	 & 61.8 & 33.6 & \textbf{28.2} & <0.001 &	 & 83.4 & 52.0 & \textbf{31.4} & <0.001 &	 & 78.6 & 75.6 & 3.0 & 0.162  \\
                        & \mixtral 8x7B Instr. 	 & 85.4 & 71.6 & \textbf{13.8} & <0.001 &	 & 98.2 & 83.8 & \textbf{14.4} & <0.001 &	 & 97.0 & 94.6 & 2.4 & 0.014  \\
                        & \gpt-3.5 Turbo & 89.2 & 87.8 & 1.4 & 0.380 &	 &  97.0 & 93.0  & \textbf{4.0} & 0.003 &  &  97.8 & 98.2  & -0.4 & 0.580  \\
                        & \gpt-4 Turbo & 90.4 & 72.4 & \textbf{18.0} & <0.001 &  &  99.2 & 91.4  & \textbf{7.8} &  <0.001 &  &  99.6 & 99.6  & 0.0 &  - \\
 \bottomrule[0.1em]
\end{tabularx}
}
    \caption{Accuracy, conditional average treatment effect (CATE), and statistical significance (p-value) on math word problems generated for the three tests detailed in \cref{sec:consistency-hypothesis}, \cref{sec:concept-test} and \cref{sec:numerical-tests}. `Co' denotes consistent, `InCo' inconsistent, `T' transfer, `C' comparison, `Ca' carry, and `NCa' no carry conditions. Results are separated by whether the prompting method is direct or chain-of-thought (CoT). `Chat' and `Inst.' indicate the instruction-tuned versions of the models. CATE values are \textbf{bolded} when significant, controlling for false discovery rate at level $\alpha{=}0.05$. Results from an additional prompting strategy, child-persona prompting, are presented in \cref{table:child_persona_results}.
    }
    \label{table:results}
    \vspace{-10pt}
\end{table*}

For each of the tests described below we select a specific feature $X$ that is localized to one of the levels, to the extent possible. That is, varying $X$ associated with a particular level should have \emph{no} effect on the levels above, and \emph{minimal} effect on the levels below. For instance, in \cref{sec:linguistic-tests} we vary the problem text without affecting the mental model, arithmetic expression or answer.%

Having selected $X$, we adapt the pipeline (\cref{sec:generation-pipeline}) to generate example pairs, one with $X=x$ and one with $X=x'$. Next, we evaluate the data quality using the procedure described in \cref{sec:data-quality}. After quality assurance, we generate a larger sample of 400 additional problem pairs, which (including the quality evaluation set) gives a total of 500 pairs for the tests. We then generate outcomes $Y(x)$ and $Y(x')$ for each of the pairs for a set of selected LLMs. We use \llama \citep{touvron2023llama} with 7B, 13B and 70B
parameters, \mistral 7B \citep{jiang2023mistral} and \mixtral 8x7B \citep{jiang2024mixtral}, \gpt-3.5 Turbo, and \gpt-4 Turbo \citep{openai2024gpt4}.
We consider both the pretrained-only and instruction-tuned versions for the \llama, \mistral and \mixtral models.
We carry out zero-shot inference with a standard prompt, a chain-of-thought prompt \citep[CoT;][]{wei_chain_of_thought}, as well as a modified ``child-persona'' CoT prompt, whose results were similar to those of the CoT setup and are presented in \cref{sec:child_persona_results}.
With the former, the models are prompted to \textit{directly} provide an answer after the input. Following previous work \cite{kojima2022large, yang2024large}, we use the format ``\textit{Q:} \texttt{\{problem\}\textbackslash n}\textit{A: The answer (Arabic numerals) is }'' for base models and ``\texttt{\{problem\}\textbackslash n}\textit{The answer (Arabic numerals) is}'' for instruction-tuned models. Then, the model prediction is retrieved from the response by extracting the first number in the model's output. 
For the CoT experimental procedure, we follow the exact method from \citet{kojima2022large}. First, the model is prompted to generate a reasoning chain by appending \textit{``Let's think step by step''} to the input. Then, the model is re-prompted to generate the final answer, which is extracted from the output as in the direct case. Responses are generated with greedy decoding and a maximum length of $256$ tokens. 
After obtaining the model's predictions, we estimate the CATE and perform a two-tailed paired sample $t$-test to determine whether the CATE estimate is significantly different from $0$. More specifically, the null hypothesis is that the two groups of model accuracy have the same mean.
We control the false discovery rate at level $\alpha{=}0.05$ using \citeposs{benjamini-hochberg-1995} procedure, considering all null hypotheses under the same bias as one distinct family.
\looseness=-1

\subsection{Problem Text: Consistency Bias}
\label{sec:linguistic-tests}\label{sec:consistency-hypothesis}
Varying the linguistic form of an otherwise equivalent problem structure can have a large effect on child performance \citep{CUMMINS1988405}. %
We test whether comparison problems with inconsistent statements are harder for LLMs than the same problems with an analogous consistent statement.\looseness=-1

\paragraph{Method.} 
Following \citeposs{Lewis1987StudentsMO} study, we consider consistent/inconsistent problem pairs where the required operation is either addition, subtraction, multiplication or division. The generation pipeline is tuned so that the problem structures follow the specification:
\vspace{1mm}
\resizebox{\linewidth}{!}{
  \begin{minipage}{\linewidth}
\begin{align*}
    & \small \container \circ \underbrace{(\transfer \mid\rate) \circ \dots \circ (\transfer \mid \rate)}_{0-2\ \text{times}} \circ \\
    & \small \comparison \circ \underbrace{(\transfer \mid \rate) \circ \dots \circ (\transfer \mid\rate)}_{0-2\ \text{times}};
\end{align*}
\end{minipage}
}
\vspace{1mm}

Note that the problems may have between $1{-}5$ reasoning steps---one for every non-container predicate. 
Apart from the first predicate (\container), only the comparison predicate introduces a new agent. The question queries the agent that was introduced by \comparison.\footnote{The following is a (consistent) example from our dataset that follows the pattern ${\small \container \circ \transfer \circ \comparison}$: ``\textit{Avery has 15 desks. Avery bought 18 desks. Natalie has 16 fewer desks than Avery. How many desks does Natalie have?}''} 
The pairs are generated such that the only sentence that varies is that which corresponds to \comparison, one being consistent and the other being its analogous inconsistent form.

\paragraph{Results.}
The results of the consistency bias test reveal 20 out of 23 statistically significant CATEs.
As displayed in \cref{table:results} (leftmost column), all models exhibit lower accuracy when solving inconsistent problems compared to their consistent counterparts. 
Interestingly, the bias appears to be exacerbated by CoT prompting, which improves the overall model performance, but 
magnifies the difference in accuracy between consistent and inconsistent problems.
This finding aligns with research indicating that CoT prompting may also amplify other types of biases present in the training data \cite{shaikh-etal-2023-second}.
Particularly notable CATEs are observed for the base versions of \llama 7B, \llama 13B, and \mistral 7B, for which the inconsistent formulation of the problems leads to an accuracy drop larger than 50\%.

\subsection{
Mental Model: Transfer vs Comparison Bias
}
\label{sec:structural-tests}
\label{sec:concept-test}
Another factor behind performance is 
that it might be harder to perform solution planning based on some mental models compared to others. We test whether LLMs are better at solving transfer-type problems than comparison problems.

\paragraph{Method.} 
The problem structures take the following forms:
\begin{align*}
    & \small \container \circ \underbrace{\transfer \circ \dots \circ \transfer}_{1-5\ \text{times}}; \\
    & \small \small \container \circ \underbrace{\comparison \circ \dots \circ \comparison}_{1-5\ \text{times}};
\end{align*}
The two problems have identical arithmetic expressions. Each comparison predicate corresponds to a comparison of a new agent with the agent introduced in the preceding sentence. Each transfer statement follows the same agent, who was introduced in the first sentence and whose state is updated through a transfer with some other agent. The problems resemble each other in linguistic form as much as possible. In particular, we make sure that the same agent names are introduced in each sentence across the two problems, in order to account for variance in performance stemming from such choices \citep{goodarzi-etal-2023-robustness}.
Consistent or inconsistent forms of comparison are sampled uniformly at random.\looseness=-1

\paragraph{Results.}
The experimental results (middle column in \cref{table:results}) show that models are consistently more accurate on problems based on transfers rather than comparisons. With the exception of CoT-prompted pretrained-only \mistral and \mixtral, we observe statistically significant positive CATEs, mirroring biases seen in children's problem-solving (\cref{sec:hypotheses}).\looseness=-1

Further, we note that the instruction-tuned models overall exhibit larger effects than pretrained-only models in the CoT setting, but not in the direct setting. This seems to be the case for the consistency bias as well. Finally, in \cref{sec:results-number-of-steps} we show some of the results as broken down by number of reasoning steps in the CoT setting.

\subsection{Arithmetic Expressions: The Carry Effect
}
\label{sec:numerical-tests}
While much of a child's performance on word problems can be explained by properties introduced by the text format, a large portion still depends on the nature of the arithmetic expression \citep{daroczy-et-al-2015-word}. We test whether LLMs are sensitive to the presence of arithmetic carry when posed with addition and subtraction in math word problems.

\paragraph{Method.} We generate pairs under the comparison specification from \cref{sec:concept-test}, but with only one step:\footnote{The one-step case follows the setups from studies on humans (\cref{sec:hypotheses}). We discovered in the previous two tests that the models frequently fail on comparison problems with only one step (which can be inferred from \cref{fig:cate_step_breakdown,fig:consistency_cot_step_breakdown}), so if a carry effect is present, it should be observable in such a setting.}
\begin{align*}
& \small \container \circ \comparison;
\end{align*}
\noindent As in \cref{sec:concept-test}, we use additive comparisons, which ensures that the arithmetic expressions only have addition and/or subtraction operators. The two problems of a pair are identical apart from the numbers. 
Following \citet{Fuerst2000carry}, we ensure that both operands as well as the answer of the problem are three-digit numbers (since children appear to rely on long-term memory for problems with small numbers; \cref{fn:memory}). One of the problems has no carry, the other has at least one (i.e., unit or tens carry).
The correct answer of the two problems is controlled to be the same.

\paragraph{Results.}
The results (rightmost column in \cref{table:results}) depart from the findings above, which gave evidence for the presence of child-like biases. In this case, model performance is similar in problems with and without carry operations---we only observe one significant result out of the 23 tests. Thus, the models seem not to be sensitive to variations isolated to the arithmetic expression level.

\section{Why do Language Models Exhibit Biases?}

A natural set of questions that arise from our results is why some child-like biases are present in the models, and why some of them (like the carry effect) are absent. The most plausible explanation in our view is the influence from the training data:
If the training data contains many examples of humans writing and solving word problems, then it may be that LLMs simulate human biases present in such text.
For instance, it may be that the distribution of the training data is skewed towards consistent problem formulations of comparison relations, which in turn could be a product of consistency bias in the humans who wrote the word problems. This would seem plausible given that the consistency bias is present also in adults \citep{Lewis1987StudentsMO, Hegarty1995ComprehensionOA}. 

Unfortunately, we cannot directly verify this hypothesis since it is unknown what data has been used to train the models considered in this study. However, as a proxy, we performed an analysis of a set consisting of word problems from  MAWPS \citep{koncel-kedziorski-etal-2016-mawps}, ASDIV-A \citep{miao-etal-2020-diverse}, and SVAMP \citep{patel-etal-2021-nlp}. These three datasets are well known and publicly available, and are thus likely to have been present in the training data of the LLMs used in this work. We found that these datasets indeed have more consistent formulations than inconsistent ones, and more transfer problems than comparison problems. More specifically, the ratios observed were 5:1 (15 and 3 in absolute numbers) and 130:9 (260 and 18 in absolute numbers), respectively.\footnote{We used the world model annotations from \citet{opedal-etal-2023-world} for this analysis, which enabled us to extract the problems with the relevant concepts. The problems containing a comparison predicate were few enough for manual inspection. For the transfer problems, $433$ in total, we manually inspected a sample of $100$. Out of these, $60$ had at least one sentence with a transfer of the same structure we consider in this work. Maximum likelihood estimation then yields $0.6\times 433 \approx 260$ relevant transfer problems. } In other words, the imbalance in problem types on these datasets is consistent with the biases we found in our analyses on text comprehension (\cref{sec:consistency-hypothesis}) and solution planning (\cref{sec:concept-test}). 

Extrapolating this hypothesis to the absence of a carry effect would imply that there is little to no difference in frequency between problems with and without carry in the training data. 
This would be harder to verify, as there are many potential traces of carry in the data beyond word problems.
Furthermore, the carry effect results suggest that there are algorithmic differences in how LLMs and humans perform arithmetic computations. In particular, the carry effect in humans is partially attributed to working memory limitations \citep{HITCH1978302}, which LLMs may not implement in the same manner. The memory and computational mechanisms through which models perform arithmetic \citep{nanda2023progress,stolfo-etal-2023-mechanistic, quirke2024understanding} are likely not affected by the increased cognitive load that the carry operation introduces in humans. 
This leads to an alternative, albeit arguably less plausible view, on why we observe the other two biases: It might be that there is algorithmic similarity between humans and language models on text comprehension and solution planning. Assessing this hypothesis would require knowledge about the mechanisms of human and language-model reasoning alike, both of which are beyond the scope of the present study. However, our results at least suggest a direction, namely, that there is at least the \emph{possibility} that the algorithms in text comprehension and solution planning exhibit similarity.

\section{Related Work
}
\label{sec:related-work}
Our work relates most closely to studies that have compared human and LLM biases on syllogisms \citep{ando-etal-2023-evaluating, eisape2023systematic}, code generation \citep{jones2022capturing}, and other non-mathematical inference tasks \citep{dasgupta2023language}. Their findings indicate that LLMs are susceptible to some of the same biases as humans, like content effects \citep{ando-etal-2023-evaluating, dasgupta2023language} and premise ordering effects \citep{eisape2023systematic}. We observe similar results in a mathematical problem-solving setting for consistency bias and transfer vs comparison bias, but not for the carry effect which relates to the step of the cognitive process that involves solving arithmetic equations.
Our study also differs from those referenced above in that we systematically compare the effect of CoT prompting to direct prompting, observing amplified effects in the CoT setting in most cases where effects are present.

We are unaware of any other work that studies cognitive biases in LLMs that, like the carry effect, relate directly to numbers. However, there seem to be similarities between the numeric representations in LLMs and the ``mental number line'' in humans \citep{shah-etal-2023-numeric}. Other studies find evidence that LLMs to some extent rely on spurious correlations in numerical reasoning \citep{razeghi-etal-2022-impact, stolfo-etal-2023-causal}, and that their performance decreases with increasing number size \citep{dziri2023faith,shen2023positional}. Beyond numerical reasoning, LLMs appear to have difficulties with causal reasoning \citep{Binz_2023, jin2023cladder,jin-et-al-2024-causation} and proof planning \citep{SaparovHe2023}.

\section{Conclusion and Implications}
This study explored whether LLMs exhibit child-like cognitive biases in arithmetic word problem-solving. We found that LLMs demonstrate biases in text comprehension and solution planning that mirror human tendencies.
Specifically, models performed better on problems where the relational keyword is consistent with the appropriate arithmetic operator compared to problems where it is not,
as well as on problems with a dynamic change of state compared to problems with a static comparison. However, at the solution execution step, LLMs did not exhibit the child-like carry effect for arithmetic computations. In general, studying biases that are present in children but not in adults may enable the disentanglement of the influence of training data from other factors that might explain language model behavior, since one would expect the training set to be heavily biased towards adult (rather than child) thinking. We therefore believe it might be a promising direction forward in language model interpretability work.

\section*{Impact Statement}

Cognitive modeling enables human simulations in place of data collection that might otherwise be unethical, harmful or costly. 
On the other hand, issues could arise if those simulations are unfaithful to human behavior.
As a broader implication of our work, we encourage practitioners to exercise care when developing and deploying cognitive models of students using LLMs, particularly, in how the student model treats numbers and other properties of arithmetic expressions. We hope that our results can provide insights for practitioners seeking to develop automated learning agents, for instance, under the tutor-learning paradigm in which a student learns by correcting the mistakes made by a computer model \citep{OKITA2014257, shahriar-2021-clarify, schmucker2023ruffle}.
We do not see any notable ethical issues with our work. 

\section*{Limitations}
\label{sec:limitations}

We cannot draw any parallels on the absolute performance in comparison with children, only on the presence or absence of each effect. This is because the datasets used in the learning science studies were not available to us. The one exception was the data from \citet{Hegarty1995ComprehensionOA}, which we evaluate on in \cref{sec:human-data}. Moreover, we do not consider the grade level of the problems, but see \citet{jiao2023automatic} for a generation method that does.

In selecting specific cognitive biases to study, we chose biases that are well-established in literature on human children and whose effects could be clearly associated with one of the steps of \cref{fig:solving-pyramid}. Another factor that fulfills these desiderata is the effect of explicit verbal cues \citep{Hudson1983CorrespondencesAN, vicente-etal-2007-influence}.
More fundamentally, a complete comparison of the biases between LLMs and humans would need to study biases that have been found in LLMs but are not necessarily present in humans. We do not take that direction into account, but we note that the number frequency effect reported by \citet{razeghi-etal-2022-impact} might be one such example.\looseness=-1

We did not use in-context examples in our evaluation since the addition of such may influence the results in ways that can be difficult to foresee. However, an interesting direction for future work could be to study whether cognitive biases can be controlled through specific choices of in-context examples or other prompts.

We stress that the conceptualization in \cref{fig:solving-pyramid} is a simplified model of the solving process.
For instance, it fails to account for shortcut strategies (see \cref{fn:shortcuts}) and it does not consider any propositional text-base representation which precedes the mental model representation in some other models \citep{Kintsch1985-KINUAS, Hegarty1995ComprehensionOA}.
We do not make any claims on the ability of LLMs to ``construct mental models'' in this work, although our results could potentially have such implications as was pointed out by a reviewer. See \cref{sec:mental-model-building} for a brief discussion. 

Finally and importantly, we only consider problems formulated in English. We note that some effects could vary across languages. For instance, the carry effect is more pronounced in German and other languages where the spelled-out order of tens and units is inverted in relation to Arabic numerical notation \citep{GOBEL201417}. Our generation pipeline can be straightforwardly adapted to other languages, and future work might consider doing so.

\section*{Acknowledgements}
We thank Emo Welzl, Ethan Gotlieb Wilcox, Julia Chatain and Yilmazcan Ozyurt for valuable feedback and discussions. Andreas Opedal acknowledges funding from the Max Planck ETH Center for Learning Systems. Alessandro Stolfo is supported by armasuisse Science and Technology through a CYD Doctoral Fellowship. 

\bibliography{refs/anthology, refs/custom}
\bibliographystyle{acl_natbib}

\newpage

\appendix

\section{More Details on the Generation Method}

\subsection{Details on Linguistic Error Correction}
\label{app:error-correction}
We use GPT-3.5 Turbo to carry out the linguistic error correction step detailed in \cref{sec:generation-pipeline}.
In \cref{table:linguistic-correction-prompt}, we provide the exact prompt used for the task. The corrected problem is generated using greedy decoding (\texttt{temperature=0}). We carry out additional integrity checks of the generated problem against the original templated text. In particular, we verify that the sentence count and relational terms (such as \textit{``more''}) are consistent post error-correction. The problem is discarded if these additional checks are not satisfied.

\begin{table}
\resizebox{\columnwidth}{!}{
    \begin{tabularx}{1\columnwidth}{X}
    \toprule[0.1em]
    Correct all grammatical mistakes that appear in the following math word problem: \texttt{[templated text]} \\
    Fix any awkward or redundant phrasing. Pay close attention to incorrect plural forms. Do NOT solve the problem. Do NOT compute any intermediate solutions. Do NOT make any changes to the numerical values or implied mathematical operations. Only output the corrected math word problem and nothing else. Do NOT restate the original problem. Do NOT include "Corrected Version:" or any description of the task. \\
 \bottomrule[0.1em]
\end{tabularx}
}
    \caption{ Prompt used for the linguistic error correction step in our generation pipeline from \cref{sec:generation-pipeline}.
    }
    \label{table:linguistic-correction-prompt}
\end{table}

\subsection{Data Quality Evaluation}
\label{sec:data-quality}

We describe the manual evaluation of the datasets generated in \cref{sec:experiments}.
For \emph{each} of the datasets, we do the following: First, generate a control set of 10 examples. These 10 examples are evaluated independently by three of this paper's authors. If there are any errors we make appropriate modifications to the pipeline and restart the procedure. If not, we proceed to evaluate 90 more examples, allocating 30 to each of the three authors. Error rate is estimated on this sample of 100 examples.

We use two binary evaluation criteria, one assessing the linguistic error correction step (iv) and one assessing test-specific attributes. A problem is deemed to be good according to the former if the generated problem only deviates from the templated text through spelling or grammar correction. The test-specific criterion and the obtained error estimates are given below.%

\paragraph{Consistency bias (\cref{sec:consistency-hypothesis}).} We evaluated data quality according to whether the two comparison statements actually were consistent and inconsistent forms to express the same comparison relationship. To be precise, the first problem needs to have a consistent relational statement for the comparison predicate, the second problem needs to have an equivalent inconsistent relational statement for the same comparison predicate, and the two problems need to be identical otherwise. Our pipeline achieved a 0\% error rate on both criteria on the 100 evaluated example problems.

\paragraph{Transfer vs comparison bias (\cref{sec:concept-test}).} We evaluated data quality according to whether the comparison and transfer problems had the same arithmetic expression and whether they followed the specified problem structure. We also ensured that the agent names and other properties matched. Our pipeline achieved a 0\% error rate on both  criteria on the 100 evaluated example problems.

\newcolumntype{Y}{>{\centering\arraybackslash}X}
\begin{table*}[t]
\resizebox{\textwidth}{!}{
    \begin{tabularx}{1.35\textwidth}{l l *{4}{Y} c *{4}{Y} c *{4}{Y}}
    \toprule[0.1em]
     \multirow{4}{*}{\textbf{Mode}} &
     \multirow{4}{*}{\textbf{Model}} &
    \multicolumn{4}{c}{\textbf{Consistency bias (\cref{sec:consistency-hypothesis})}} & &
    \multicolumn{4}{c}{\textbf{Transfer vs comparison bias (\cref{sec:concept-test})}} & & 
    \multicolumn{4}{c}{\textbf{Carry effect (\cref{sec:numerical-tests})}} \\\cmidrule{3-6}\cmidrule{8-11}\cmidrule{13-16}
     &  & \multicolumn{3}{c}{\textbf{Accuracy (\%)}}& \multirow{2}{*}{$p$-value} & &
     \multicolumn{3}{c}{\textbf{Accuracy (\%)}}& \multirow{2}{*}{$p$-value} & &
     \multicolumn{3}{c}{\textbf{Accuracy (\%)}}& \multirow{2}{*}{$p$-value} \\\cmidrule{3-5}\cmidrule{8-10}\cmidrule{13-15}
      & & \textbf{Co} & \textbf{InCo}  &  CATE & &
     & \textbf{T} & \textbf{C}  &  CATE & & 
     & \textbf{NCa} & \textbf{Ca}   &  CATE &\\
    \midrule
     \multirow{8}{*}{Child CoT} & \llama 7B 	 & 14.6 & 5.0 & \textbf{9.6} & <0.001 &	 & 19.8 & 11.6 & \textbf{8.2} & <0.001 &	 & 40.0 & 44.4 & -4.4 & 0.048  \\
& \llama 13B 	 & 20.0 & 5.0 & \textbf{15.0} & <0.001 &	 & 35.0 & 7.0 & \textbf{28.0} & <0.001 &	 & 20.8 & 21.0 & -0.2 & 0.903  \\
& \mistral 7B 	 & 39.2 & 17.2 & \textbf{22.0} & <0.001 &	 & 48.8 & 26.6 & \textbf{22.2} & <0.001 &	 & 58.4 & 57.2 & 1.2 & 0.415  \\
& \mixtral 8x7B 	 & 66.2 & 34.6 & \textbf{31.6} & <0.001 &	 & 69.8 & 49.4 & \textbf{20.4} & <0.001 &	 & 70.4 & 69.4 & 1.0 & 0.701  \\
                                     \cmidrule{2-16} 
& \llama 7B Chat 	 & 55.2 & 24.2 & \textbf{31.0} & <0.001 &	 & 62.6 & 33.8 & \textbf{28.8} & <0.001 &	 & 67.8 & 63.0 & 4.8 & 0.069  \\
& \llama 13B Chat 	 & 65.2 & 27.0 & \textbf{38.2} & <0.001 &	 & 79.8 & 48.2 & \textbf{31.6} & <0.001 &	 & 80.0 & 77.0 & 3.0 & 0.108  \\
& \mistral 7B Instr. 	 & 65.0 & 30.6 & \textbf{34.4} & <0.001 &	 & 75.2 & 52.8 & \textbf{22.4} & <0.001 &	 & 77.2 & 74.4 & 2.8 & 0.178  \\
& \mixtral 8x7B Instr. 	 & 88.6 & 72.4 & \textbf{16.2} & <0.001 &	 & 98.8 & 82.4 & \textbf{16.4} & <0.001 &	 & 97.6 & 97.4 & 0.2 & 0.809  \\
 \bottomrule[0.1em]
\end{tabularx}
}
    \caption{Accuracy, conditional average treatment effect (CATE), and statistical significance ($p$-value) on math word problems generated for the three tests detailed in \cref{sec:consistency-hypothesis}, \cref{sec:concept-test} and \cref{sec:numerical-tests}. `Co' denotes consistent, `InCo' inconsistent, `T' transfer, `C' comparison, `Ca' carry, and `NCa' no carry conditions. The results presented are for the child-persona prompting strategy described in \cref{sec:child_persona_results}. `Chat' and `Inst.' indicate the instruction-tuned versions of the models. CATE values are \textbf{bolded} when $p<0.001$.
    }\label{table:child_persona_results}
\end{table*}

 \vspace{-25pt}

\begin{figure}[t]
    \centering
\includegraphics[width=0.47\textwidth]{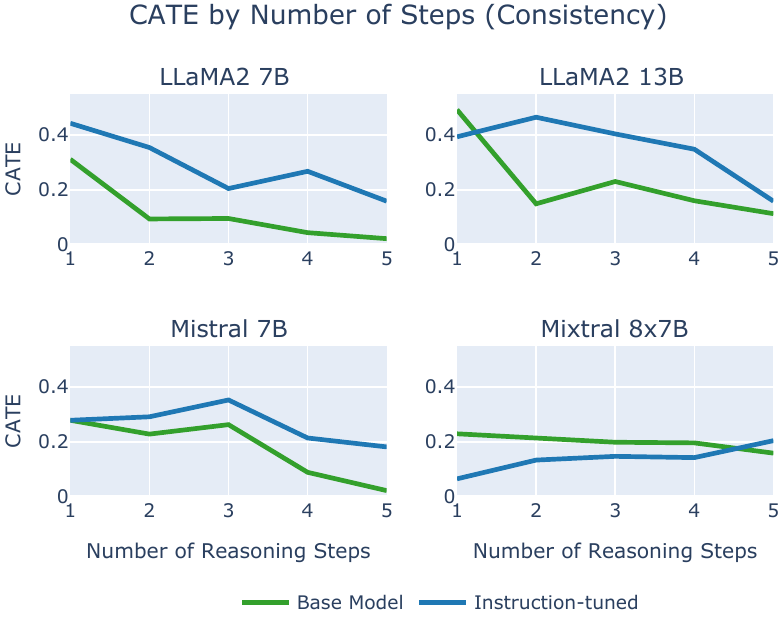}
    \caption{CATE for the consistency effect (\cref{sec:consistency-hypothesis}) in  CoT-prompted models stratified by number of reasoning steps. `Instruction-tuned' refers to the \textit{Chat} and \textit{Instruct} versions of \llama and \mistral/\mixtral, respectively. }
    \label{fig:consistency_cot_step_breakdown}
\end{figure}

\begin{figure}[t]
    \centering
    \includegraphics[width=0.47\textwidth]{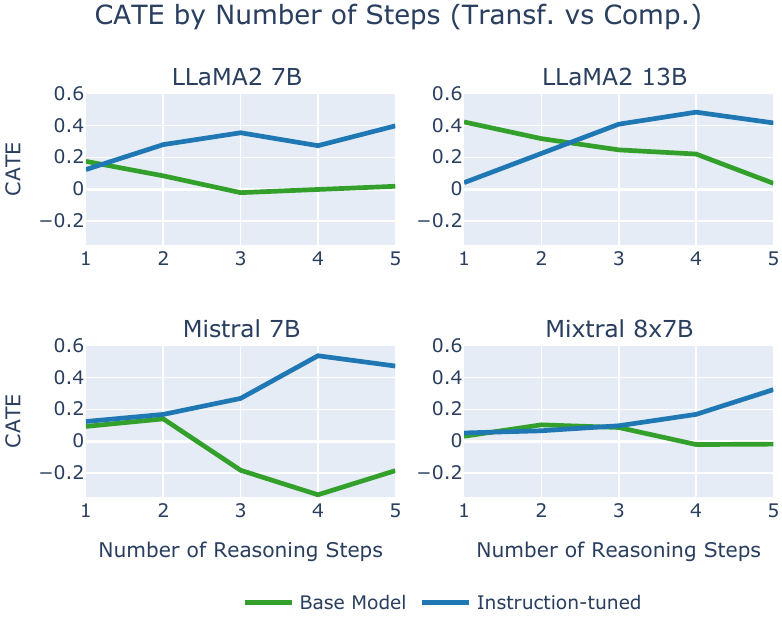}
    \vspace{-8pt}
    \caption{CATE for transfer vs comparison (\cref{sec:concept-test}) in the CoT-prompted case stratified by
    number of reasoning steps. `Instruction-tuned' refers to the \textit{Chat} and \textit{Instruct} versions of \llama and \mistral/\mixtral, respectively. Base and instruction-tuned models display opposite relationships between length and bias strength.
}
    \label{fig:cate_step_breakdown}
\end{figure}

\paragraph{The carry effect (\cref{sec:numerical-tests}).} We evaluated data quality according to whether one problem had no carry computation steps, the other one had at least one, and they were equal otherwise. That is, only the numbers differed and the two problems had the same answer. Our pipeline achieved a 0\% error rate on both criteria on the 100 evaluated example problems.

\subsection{Related Methods} 
\label{sec:related-methods}
Our generation pipeline differs from \citeposs{opedal-etal-2023-world} method 
in that we generate intermediate templated texts, while they generate the problem texts directly conditioned on mental models. 
\citet{Polozov2015PersonalizedMW} also generates word problems from logical representations, but their approach does not allow explicit control over arithmetic concepts, which is an important factor underlying difficulty level; \cref{sec:hypotheses}. While our experiments require strict control over linguistic form, our error correction step could in principle be broadened to perform paraphrasing and theme rewriting \citep{koncel-kedziorski-etal-2016-theme} as well.

\section{Additional Results}\label{sec:additional-results}
\subsection{Child-Persona Prompting}
\label{sec:child_persona_results}

Taking inspiration from claims that LLMs can act as agent models \citep{andreas-2022-language}, we also experimented with an additional prompt in which the LLM is instructed to impersonate a grade-school child. Specifically, we employ a modified version of the zero-shot chain-of-thought prompt, tailored to simulate a child's reasoning process. We prompt the model with the phrase \textit{``Let’s think step by step as a grade-school child would,''} replacing the standard CoT instruction. Following this, we apply the same decoding method used in traditional CoT. The results for this approach are reported for the open-source models (apart from \llama 70B) in \cref{table:child_persona_results}. While we notice larger consistency and transfer vs comparison effects for some models, we observe no substantial departure from the results achieved with conventional CoT prompting.

\subsection{Bias Strength by Number of Reasoning Steps}
\label{sec:results-number-of-steps}

In \cref{fig:cate_step_breakdown}
we show how the CATE of the transfer vs comparison bias varies with the number of reasoning steps in the problems for the CoT setting. Interestingly, we observe that the CATE sizes increase with the number of reasoning steps for the instruction-tuned models, whereas they decrease for the pretrained-only base models. We are unaware of literature on the relationship between human transfer vs comparison bias and the number of steps, so we can not make any claims about which of these patterns is more cognitively plausible. 

The consistency-effect test does not exhibit such diverging trends for the CATEs.
\cref{fig:consistency_cot_step_breakdown} illustrates how the strength of the measured biases change in relation to the number of reasoning steps in a problem (in the CoT-prompted case). Note that the carry effect experiments were carried it out for problems with only one step.

\subsection{Data from \citet{Hegarty1995ComprehensionOA}}
\label{sec:human-data}
In \cref{table:hegarty} we present the results on a few selected models when evaluated on the data from the study by \citet{Hegarty1995ComprehensionOA}. Theirs was the only background study for which we were able to obtain the data that was used. The dataset contains eight problem pairs and targets the consistency bias. While we do not obtain significant results, we do get an indication of the absolute effect of the bias as compared to the human subjects in \citeposs{Hegarty1995ComprehensionOA} study (who were undergraduate college students). In their study, solvers who committed at least four errors out of the total of 16 problems had an average accuracy of $62\%$ and $24\%$, on consistent problems and inconsistent problems respectively. None of the absolute accuracies in \cref{table:hegarty} are similar, but \mixtral 8x7B Instruct displays a similar absolute effect size ($37.5\%$ vs $38\%$).\looseness=-1

\section{Brief Discussion on Mental Model Building}\label{sec:mental-model-building}
We note that the presence of consistency bias could be viewed as an argument against the position that language models construct something akin to a mental model during problem-solving. 
Indeed, people who exhibit consistency bias seem to be more likely to construct a mental model of the problem compared to those who do not, based on eye-fixation behavior \citep{Hegarty1995ComprehensionOA}.
Intuitively, a (human or non-human) solver that constructs a mental model should be able to be more robust against inconsistent phrasings, assuming that the text-comprehension step of the solving pipeline is not made significantly harder by inconsistent phrasings.

\begin{table}[t]
\resizebox{0.48\textwidth}{!}{
\begin{tabular}{l l l l l l}
    \toprule[0.1em]
     \multirow{4}{*}{\textbf{Mode}} &
     \multirow{4}{*}{\textbf{Model}} &
    \multicolumn{4}{c}{\textbf{Consistency bias}}  \\\cmidrule{3-6}
     &  & \multicolumn{3}{c}{\textbf{Accuracy (\%)}}& \multirow{2}{*}{$p$-value} \\\cmidrule{3-5}
      & & \textbf{Co} & \textbf{InCo}  &  CATE &  \\
    \midrule
     \multirow{3}{*}{Direct} 
& \mixtral 8x7B 	 & 0 & 0 & 0 & -  \\
& \mixtral 8x7B Instr. 	 & 12.5 & 0 & 12.5 & 0.35  \\
& \gpt-3.5 Turbo 	 & 62.5 & 62.5 & 0 & -  \\
\midrule
     \multirow{3}{*}{CoT} 
& \mixtral 8x7B 	 & 37.5 & 12.5 & 25 & 0.35  \\
& \mixtral 8x7B Instr. 	 & 100 & 62.5 & 37.5 & 0.08  \\
& \gpt-4 Turbo 	 & 75 & 62.5 & 12.5 & 0.35  \\
 \bottomrule[0.1em]
\end{tabular}
}
    \caption{Accuracy, conditional average treatment effect (CATE), and statistical significance ($p$-value) on word problems from \citet{Hegarty1995ComprehensionOA}.
    }
    \label{table:hegarty}
\end{table}

\end{document}